\documentclass[11pt,a4paper]{article}
\usepackage{times,latexsym}
\usepackage{url}
\usepackage[T1]{fontenc}
\usepackage{graphicx}
\usepackage{xcolor}
\usepackage{amsmath}
\usepackage{mathtools}
\usepackage{booktabs}

\usepackage{tikz-qtree}
\usepackage{forest}

\usepackage{xcolor}

\usepackage[]{acl}

\title{Characterizing Idioms: Conventionality and Contingency}

\author{Michaela Socolof\textsuperscript{1,2}, Jackie Chi Kit Cheung\textsuperscript{1,2,3}, Michael Wagner\textsuperscript{1}, Timothy J. O'Donnell\textsuperscript{1,2,3} \\
  McGill University\textsuperscript{1}, Quebec AI Institue, Mila\textsuperscript{2}, Canada CIFAR AI Chair\textsuperscript{3}\\
  \texttt{michaela.socolof@mail.mcgill.ca}, \texttt{chael@mcgill.ca},\\
  \texttt{jcheung@cs.mcgill.ca}, \texttt{timothy.odonnell@mcgill.ca}}

\begin{document}
\maketitle
\begin{abstract}
Idioms are unlike most phrases in two important ways. First, words in an idiom have non-canonical meanings. Second, the non-canonical meanings of words in an idiom are contingent on the presence of other words in the idiom. Linguistic theories differ on whether these properties depend on one another, as well as whether special theoretical machinery is needed to accommodate idioms. We define  two measures that correspond to the properties above, and we implement them using BERT \citep{bert} and XLNet \citep{xlnet}. We show that English idioms fall at the expected intersection of the two dimensions, but that the dimensions themselves are not correlated. Our results suggest that special machinery to handle idioms may not be warranted. %We also find that the syntactic head word of an idiom tends to have a more standard meaning than the non-head elements.
\end{abstract}

\section{Introduction}

Idioms---expressions like \textit{rock the boat}---bring together two
phenomena which are of fundamental interest in understanding
language. First, they exemplify \textit{non-conventional word
  meaning} \citep[]{Weinreich:1969, Nunberg:1994}. The words
\textit{rock} and \textit{boat} in this idiom seem to carry
particular meanings---something like \textit{destabilize} and
\textit{situation}, respectively---which are different from the
conventional meanings of these words in other contexts. Second, unlike
other kinds of non-conventional word use such as novel metaphor, there
is a contingency relationship between words in an idiom
\citep[]{wood1986definition, Pulman:1993}. It is the specific
combination of the words \textit{rock} and \textit{boat} that has
come to carry the idiomatic meaning. \textit{Shake the canoe} does not have the same accepted meaning.
  
In the literature, most discussions of idioms make use of prototypical
examples such as \textit{rock the boat}. This obscures an
important fact: There is no generally agreed-upon definition of
\textit{idiom}; phrase types such as light verb constructions (e.g.,
\textit{take a walk}) and semantically transparent collocations
(e.g., \textit{now or never}) are sometimes included in the class \citep[e.g.,][]{Palmer:1981} and
sometimes not \citep[e.g.,][]{Cowie:1981}. 
This lack of homogeneity among idiomatic phrases has been recognized as a challenge in the domain of NLP, with \citet{sag} suggesting that a variety of techniques are needed to deal with different kinds of multi-word expressions.
What does
seem clear is that prototypical cases of idiomatic phrases tend to
have higher levels of both non-conventional meaning and contingency
between words.

This combination of non-conventionality and contingency has led to a
number of theories that treat idioms as exceptions to the mechanisms
that build phrases compositionally. These theories posit special
machinery for handling idioms \citep[e.g.,][]{Weinreich:1969,
  BobrowBell:1973, SwinneyCutler:1979}. An early but representative
example of this position is \citet{Weinreich:1969}, who posits the
addition of two structures to linguistic theory: (1) an \textit{idiom
  list}, where each entry contains a string of morphemes, its
associated syntactic structure, and its sense description, and (2) an
\textit{idiom comparison
  rule}, which matches strings against the idiom list. Such theories
must of course provide principles for addressing the difficult
problem of distinguishing idioms from other instances of
non-conventionality or contingency.
  
We propose an alternative approach, which views idioms not as
exceptional, but merely the result of the interaction of two
independently motivated cognitive mechanisms. The
first allows words to be interpreted in non-canonical ways
depending on context. The second
allows for the storage and reuse of linguistic structures---not just words, but larger phrases as well
\citep[e.g.,][]{DiSciullo:1987, Jackendoff:2002, O'Donnell:2015}. There is disagreement in the literature
about the relationship between these two properties; some theories of representation
predict that the only elements that get stored are those with
non-canonical meanings \citep[e.g.,][]{Bloomfield:1933,
  PinkerPrince:1988}, whereas others predict that storage can happen
no matter what \citep[e.g.,][]{O'Donnell:2015, TremblayBaayen:2010}.
We predict that, consistent with the latter set of theories, neither mechanism should depend on the
other.%, and further, we expect to find linguistic constructions throughout the space where the mechanisms interact.

This paper presents evidence that prototypical idioms occupy a
particular region of the space of these two mechanisms, but are not
otherwise exceptional. We define two measures,
\textit{conventionality}---meant to measure the degree to which words
are interpreted in a canonical way, and \textit{contingency}---a statistical association measure meant
to capture the degree to which the presence of one word form depends on the
presence of another. Our implementations make use of the pre-trained language models BERT \citep{bert} and XLNet \citep{xlnet}. We construct a novel corpus of English phrases typically
called idioms, and show that these phrases fall at the intersection of
low conventionality and high contingency, but that the two measures
are not correlated and there are no clear discontinuities that
separate idioms from other types of phrases.

Our experiments also reveal hitherto unnoticed asymmetries in the
behavior of head and non-head words of idioms. In idioms, the
dependent word (e.g., \textit{boat} in \textit{rock the boat})
shows greater deviation from its conventional meaning than the head.
% We find that coordinate structures (e.g. \textit{X and Y}) pattern consistently with other syntactic structures in this respect, as long as the first conjunct is considered the head. This may shed light on the longstanding debate over the headedness of coordinate structures.

%In what follows, we present the background work that informs our study, plus descriptions of our methods, detailing the creation of our dataset and the measures we use to calculate conventionality and contingency. This is followed by the results of our three experiments and discussion of the results.

\section{Conventionality and contingency}

%\fix{There are various positions in the linguistics literature about the relationship between compositionality and storage. On one hand, there are theories predicting that the only elements that get stored are those that are non-compositional \citep[e.g.,][]{Bloomfield:1933, PinkerPrince:1988}. On the other hand, some theories predict that storage can happen regardless of whether something is compositional \citep[e.g.,][]{O'Donnell:2015, TremblayBaayen:2010}.}

%\fix{To explore the relationship between compositionality and storage, we present measures that are intended to capture these two phenomena, and we probe for evidence of a dependence between storage and non-compositionality.}
In this section we describe the motivation behind our two measures and
lay out our predictions about their interaction.

Our first measure, \textit{conventionality}, captures the extent to
which subparts of a phrase contribute their normal meaning to the
phrase. Most of language is highly conventional; we can combine a
relatively small set of units in novel ways, precisely because we can
trust that those units will have similar meanings across contexts. At
the same time, the linguistic system allows structures like
metaphors and idioms, which use words in non-conventional ways. Our
conventionality measure is intended to distinguish phrases based on
how conventional the meanings of their words are.

Our second measure, \textit{contingency}, captures how unexpectedly
often a group of words occurs together in a phrase and, thus, measures
the degree to which there is a statistical contingency---the presence
of one or more words strongly signals the likely presence of the others.
This notion of contingency has also been argued to be a critical
piece of evidence used by language learners in deciding which
linguistic structures to store \citep[e.g.,][]{hay.j:2003,O'Donnell:2015}.

To aid in visualizing the space of phrase types we expect to find in
language, we place our two dimensions on the axes of a 2x2
matrix, where each cell contains phrases that are either high or low
on the conventionality scale, and high or low on the contingency
scale. The matrix is given in Figure~\ref{table:2by2},
with the types of phrases we expect in each cell.

%\hspace{2cm}
\begin{figure}[htb]
\begin{centering}
\scalebox{0.75}{
\begin{tabular}{|c|c|c|}
    \hline
     & \textbf{Low} & \textbf{High} \\
     & \textbf{conv.} & \textbf{conv.}\\
    \hline
    \textbf{High} & Idioms & Common \\
    \textbf{cont.} & \textit{(e.g., raise hell)} & collocations \\
    & & \textit{(e.g., in and out)} \\
    \hline
    \textbf{Low} & Novel & Regular \\
    \textbf{cont.} & metaphors & language use\\
    & & \textit{(e.g., eat peas)} \\
    \hline
\end{tabular}}
\caption{Matrix of phrase types, organized by whether they have
  high/low conventionality and high/low contingency}
\label{table:2by2}
\end{centering}
\end{figure}

We expect our measures to place idioms primarily in the top left
corner of the space. At the same time, we predict a lack of
correlation between the measures and a lack of major discontinuities
in the space. We take these predictions to be consistent with theories
that factorize the problem into two mechanisms (captured by our
dimensions of conventionality and contingency). We contend that
this factorization provides a natural way of characterizing
not just idioms, but also collocations and novel metaphors, alongside
regular language use.

%\tjo{maybe soften this language?}  Our findings also have implications
%for theories of meaning. Specifically, the results suggest that
%so-called compositionality exists on a cline, providing evidence for
%the idea that words can contribute partial information about their
%meaning to the larger phrases they are part of, and that this is a
%robust part of the linguistic system.

%\tjo{so I think that this section could go after the methods, to make
 % it easier to reference what we did in reverse and shorten a bit. but
 % it isn't critical.}

\section{Methods}
In this section, we describe the creation of our corpus of idioms and
define measures of conventionality and contingency. Given that definitions of idioms differ in which phrases in our
dataset count as idioms (some would include semantically transparent
collocations, others would not), we do not want to commit to any
particular definition a priori, while still acknowledging that people share somewhat weak but broad intuitions about idiomaticity. As we discuss below, our idiom dataset consists of phrases that have at some point been called idioms in the linguistics literature.% Most importantly, given that we are investigating whether idioms can be characterized on dimensions that do not make reference to the notion of an idiom, we have tried to refrain from presupposing an a priori category of “idiom” in our setup, while still acknowledging that people share somewhat weak but broad intuitions about idiomaticity.

%Our experiments were designed to investigate whether the conventionality and contingency measures could distinguish idioms, as well as to explore whether the two measures were correlated with each other.\tjo{this is a weird place to have this comment, you don't need to repeat stuff like this which is sort of repetitive and also a bit vague, instead, all the text should be written with the goals you are trying to achieve for the reader in mind.}

%The experiments use a parsed corpus of phrases with their contexts. The types of phrases include idioms, collocations, and regular phrases that are not common expressions. The details of our corpus are described below. For each phrase, we calculate a conventionality score and a contingency score.

%Furthermore, we expect to see a cline on both dimensions, with phrases that lie at intermediate levels of both conventionality and contingency. This is in contrast to the \textit{words with spaces} theory, which predicts that the conventionality measure (essentially a proxy for compositionality) should pull out idioms as a distinct cluster.

\subsection{Dataset}

We built a corpus of sentences containing idioms and non-idioms, all
gathered from the British National Corpus \citep[BNC;][]{BNC}, which is a 100 million word collection of written and spoken English from the late twentieth century. The
corpus we construct is made up of sentences containing \textit{target phrases} and
\textit{matched
  phrases}, which we detail below.

The target phrases in our corpus consist of 207 English phrasal
expressions, some of which are prototypical idioms (e.g., \textit{rock
  the boat}) and some of which are boundary cases that are sometimes considered idioms, such as collocations (e.g., \textit{bits
  and pieces}). These expressions
are divided into four categories based on their syntax: verb object
(VO), adjective noun (AN), noun noun (NN), and binomial (B)
expressions. Binomial expressions are fixed pairs of words joined by
\textit{and} or \textit{or} (e.g., \textit{wear and tear}). %The breakdown of phrases into these four categories is as follows: 31 verb object pairs, 36 adjective noun pairs, 33 noun noun pairs, and 58 binomials.
The phrases were selected from lists of idioms
published in linguistics papers \citep[]{Riehemann:2001, Morgan:2016,
  Stone:2016, Brueningetal:2018, Bruening:2019, Titoneetal:2019}. We added the lists to our dataset one-by-one until we had at least 30 phrases of each syntactic type. We chose these four types in advance to investigate a variety of syntactic types to prevent our results from being too heavily skewed by any potential syntactic confounds in particular constructions. The
full list of target phrases is given in Appendix A. The numerical
distribution of phrases is given in Table~\ref{table:dataset}.

\begin{table}[htb]
\begin{centering}
\scalebox{0.9}{
\begin{tabular}{|c|c|c|}
    \hline
     \textbf{Phrase} & \textbf{Number of} & \textbf{Example}\\
     \textbf{type} & \textbf{phrases} & \\
    \hline
    \textbf{VO} & 31 & \textit{\textbf{jump} the gun} \\
    \hline
    \textbf{NN} & 36 & \textit{word \textbf{salad}} \\
    \hline
    \textbf{AN} & 33 & \textit{red \textbf{tape}}\\
    \hline
    \textbf{B} & 58 & \textit{\textbf{fast} and loose} \\
    \hline
\end{tabular}}
\caption{Types, counts, and examples of target phrases in our idiom corpus, with head words bolded}
\label{table:dataset}
\end{centering}
\end{table}

The BNC was constituency parsed using the Stanford Parser
\citep[][]{stanfordparser}, then Tregex \citep[][]{tregex}
expressions were used to find instances of each target phrase.

Matched, non-idiomatic sentences were also extracted in order to allow for direct comparison of conventionality scores for the same word in idiomatic and non-idiomatic contexts. To obtain these
matches, we used Tregex to find sentences that included a phrase with
the same syntactic structure as the target phrase. Each target phrase
was used to obtain two sets of matched phrases: one set where the head
word remained constant and one where the non-head word remained
constant.\footnote{To obtain matched phrases, we follow work such as \citet{gazdar.g:1981}, \citet{Rothstein:1991}, and \citet{Kayne:1994} in treating the first element in a binomial as the head. We discuss this further in Section~\ref{asymmetriessec}.} For example, to get head word matches of the adjective noun
combination \textit{sour grapes}, we found sentences where the lemma
\textit{grape} was modified with an adjective other than
\textit{sour}. Below is an example of a sentence found by
this method:
\\

\hfill\begin{minipage}{\dimexpr\textwidth-3cm}
\textit{Not a \textbf{special grape} for winemaking, nor}\\ 
\textit{a hidden architectural treasure, but hot}\\
\textit{steam gushing out of the earth.}
\end{minipage}
\\

%\textit{It's called simply `\textbf{White Grapes}.'}
%\\

%\textit{The last member of the party sitting at}\\
%\textit{Miss Jarman's elbow - a quietly dressed} \\
%\textit{individual with the face of Punch, but}\\
%\textit{with eyes like \textbf{wet grapes} - glanced }\\
%\textit{across the table.}
%\end{minipage}
%\\

The number of instances of the matched phrases ranged from 29 (the number of verb object phrases with the object \textit{logs} and a verb other than \textit{saw}) to the tens of thousands (e.g., for verb object phrases beginning with \textit{have}), with the majority falling in the range of a few hundred to a few thousand. Issues of sparsity were more pronounced among the target phrases, which ranged from one instance (\textit{word salad}) to 2287 (\textit{up and down}). Because of this sparsity, some of the analyses described below focus on a subset of the phrases.

The syntactic consistency between the target and matched phrases is an
important feature of our corpus, as it allows us to compare conventionality across semantic contexts while controlling for
syntactic structure.

\subsection{Conventionality measure} \label{convmeasure}

Our measure of conventionality is built on the idea that a word being
used in a conventional way should have similar or related meanings
across contexts, whereas a non-conventional word meaning can be
idiosyncratic to particular contexts. In the case of idioms, we expect
that the difference between a word’s meaning in an idiom and the
word’s conventional meaning should be large. On the other hand, there should be little difference
between the word’s meaning in a non-idiom and the word’s conventional
meaning.

Our measure makes use of the language model BERT \cite{bert} to obtain
contextualized embeddings for the words in our
dataset. BERT was trained on a corpus of English text, both nonfiction and fiction, with the objectives of masked language modeling and next sentence prediction. %BERT was trained on English Wikipedia and the BooksCorpus \cite{zhu}. Likewise, the BNC, which we are using, contains a mix of fiction and nonfiction sources.
For each of our phrases, we compute the conventionality measure
separately for the head and non-head words. For each case (head and
non-head), we first take the average embedding for the word across
sentences \textit{not containing} the phrase. That is, for \textit{rock} in \textit{rock the
  boat}, we get the embeddings for the word \textit{rock} in
sentences where it does not occur with the direct object
\textit{boat}. Let $O$ be a set of instances $w_1, w_2, ... , w_n$ of
a particular word used in contexts \textit{other than} the context of the target
phrase. Each instance has an embedding
$u_{w_1}, u_{w_2}, ... , u_{w_n}$. The average embedding for the word
among these sentences is:
\begin{equation}
\mu_O = \frac{1}{n}\sum_{i=1}^n u_{w_i}
\end{equation}

We take this quantity to be a proxy for the prototypical, or conventional, meaning of
the word. The conventionality score
is the negative of the average distance between $\mu_O$ and the embeddings for uses of
the word across instances of the phrase in question. We compute this as follows:
\begin{equation}
\mathrm{conv}(\mathrm{phrase}) = -\frac{1}{m} \sum_{i=1}^m \left \Vert \frac{T_i - \mu_O}{\sigma_O}
\right \Vert_2
\end{equation}

\noindent
where $T$ is the embedding corresponding to a particular use of the
word in the target phrase, and $\sigma_O$ is the component-wise standard
deviation of the set of embeddings $u_{w_i}$, and $m$ is the number of sentences in which the target phrase is used.

\subsection{Contingency measure}

Our second measure, which we have termed \textit{contingency}, refers
to whether a particular set of words appears within the same phrase at
an unexpectedly high rate. The measure is based on the notion of pointwise mutual information (PMI), which is a measure of the strength of association between two events. We use a
generalization of PMI that extends it to sets of more than two events,
allowing us to capture the association between phrases that contain
more than two words.

The specific generalization of PMI that we use has at various times
been called total correlation \citep{Watanabe:1960}, multi-information
\citep{Studeny:1998}, and specific correlation \citep{VandeCruys:2011}.
\begin{equation}
  \mathrm{cont}(x_{1},x_{2},...,x_{n}) = 
  \log\frac{p(x_{1},x_{2},...,x_{n})}{\prod_{i=1}^{n} p(x_{i})}
\end{equation}

For the case of three variables, we get:

\begin{equation}
  \mathrm{cont}(x,y,z) = 
  \log\frac{p(x,y,z)}{p(x)p(y)p(z)}
\end{equation}

To estimate the contingency of a phrase, we use word
probabilities given by XLNet \citep{xlnet}, an auto-regressive
language model that gives estimates for the conditional probabilities
of words given their
context. Like BERT, XLNet was trained on a mix of fiction and nonfiction data. %For a given phrase, we compute an estimate of generalized PMI using conditional probabilities of words in context, with particular words masked out.\tjo{a little confusing. maybe easier to give math then explain it?}
To estimate the joint probability of the words in \textit{rock the
  boat} in some particular context (the numerator of the expression
above), we use XLNet to obtain the product of the conditional
probabilities in the chain rule decomposition of the joint. We get the
relevant marginal probabilities by using attention masks over particular words, as
shown below, where \textit{c} refers to the context---that is, the
rest of the words in the sentence containing \textit{rock the boat}.

\[
\begin{array}{l}
\Pr(boat \mid rock\ the,\ c) = .. rock\ the\ \textbf{boat} ... \\
\Pr(the \mid rock,\ c) \hspace{1.05cm} = ... rock\ \textbf{the}\ [\_\_\_] ... \\
\Pr(rock \mid c) \hspace{1.85cm} = ... \textbf{rock}\ [\_\_\_]\ [\_\_\_] ...
\end{array}
\]

The denominator is the product of the probabilities of each individual
word in the phrase, with both of the other words masked out:

\[
\begin{array}{l}
\Pr(boat \mid c) = ... [\_\_\_]\ [\_\_\_]\ \textbf{boat} ... \\
\Pr(the \mid c) \hspace{.4cm} = ... [\_\_\_]\ \textbf{the}\ [\_\_\_] ... \\
\Pr(rock \mid c) \hspace{.2cm} = ... \textbf{rock}\ [\_\_\_]\ [\_\_\_] ... 
\end{array}
\]

The conditional probabilities were computed right to left, and included the sentence to the left and the sentence to the right of the target sentence for context. Note that in order to have an interpretable chain rule decomposition for each sequence, we calculate the XLNet-based generalized PMI for
the entire string bounded by the two words of the idiom---this means,
for example, that the phrase \textit{rock the fragile boat} will
return the PMI score for the entire phrase, adjective included.

\section{Validation of conventionality measure}\label{validation}

Our conventionality measure provides an indirect way of looking at how canonical a word's meaning is in context. In order to validate that the measure corresponds to an intuitive notion of unusual word meaning, we carried out an
online experiment to see whether human judgments of conventionality
correlated with our automatically-computed conventionality scores. The
experimental design and results are described below. (Note that our contingency measure directly computes the statistical quantity we want, so validation is not necessary.)

\subsection{Human rating experiment}

The experiment asked participants to rate the literalness of a word or
phrase in context.\footnote{Participants were recruited on Amazon Mechanical Turk and compensated at a rate of \$15/hour. The study was carried out with REB approval.} We used twenty-two verb object
target phrases and their corresponding matched phrases.\footnote{We excluded one target phrase from the analyses (\textit{spill the beans}) based on examination of the BERT-based conventionality scores. The verb \textit{spill} used in \textit{spill the beans} scored anomalously high on conventionality; investigation of the target and matched sentences revealed that roughly half of the matched sentences included a different idiom: \textit{spill X's guts}. We checked the rest of our dataset and did not find other instances of this confound.}
For each
target phrase (e.g., \textit{rock the boat}), there were ten items,
each of which consisted of the target phrase used in the context of a
(different) sentence. Each sentence was presented with the preceding
sentence and the following sentence as context, which is the same
amount of context that the automatic measure was given. In each item,
a word or phrase was highlighted, and the participant was asked to
rate the literalness of the highlighted element. We obtained judgments
of the literalness of the head word, non-head word, and entire phrase
for ten different sentences containing each target phrase.

We also obtained literalness judgments of the head word and entire
phrase for phrases matched on the head of the idiom (e.g., verb object phrases with
\textit{rock} as the verb and a noun other than \textit{boat} as
the object). Similarly, we obtained literalness judgments of the
non-head word and the entire phrase for phrases matched on the
non-head word of the idiom (e.g., verb object phrases with \textit{boat}
as the object and a verb other than \textit{rock}). Participants
were asked to rate literalness on a scale from 1 (`Not literal at
all') to 6 (`Completely literal'). We chose to use an even number of points on the scale to discourage participants from imposing a three-way partition into `low', 'neutral', and 'high'. Items were presented using a Latin
square design. The experiment was run online using the Prosodylab Experimenter \citep{prosodylab}, a JavaScript tool building on jsPsych \citep{jspsych}.

Participants were adult native English speakers who gave written
informed consent to participate. %They werecompensated at a rate of \$15/hour, and
The experiment took about 10 minutes to complete. The data were recorded using
anonymized participant codes, and none of the results included any
identifying information. There were 150 participants total. The data from 10 of those participants were excluded due to failure to follow the instructions (assessed with catch trials).
  
\subsection{Results}  
  
To explore whether our conventionality measure correlates with human
judgments of literalness, we compare the scores to the results from
the rating experiment. Ratings were between 1 and 6, with 6
being the highest level of conventionality.

      We predicted that the literalness ratings should
      increase as conventionality scores increased. To assess whether our
      prediction was borne out, a linear mixed model was fit using the lmerTest \cite{lmer}
      package in R \cite{r}, with conventionality score and highlighted
      word (head versus non-head) and their interaction as predictors, plus random effects of
      participant and item.\footnote{\scriptsize \texttt{Rating$\sim$Conv*Head$+$(1|Item)$+$(1+Conv||Partp)}} All random effects were maximal up to convergence. Results are shown in
      Table \ref{modelreport} in Appendix B. The results confirm our prediction that
      words that receive higher conventionality scores are rated as highly literal by humans ($\hat{\beta} = 0.185$, $SE(\hat{\beta}) = 0.050$, $p < 0.001$; see Row 2 of Table~\ref{modelreport} in Appendix B).

        We carried out a nested model comparison to see whether
        including the BERT conventionality score as a predictor
        significantly improved the model, and we found that it did. A likelihood ratio test with
        the above model and one without the BERT
        conventionality score as a predictor yielded a higher log
        likelihood for the full model ($\chi^2 = 80.043$, $p < 0.001$).

\section{Analyses} \label{analyses}

In this section we present analyses of our two measures individually,
showing that they capture the properties they were intended to capture. We then investigate the interaction between the measures. Section~\ref{mainevent} evaluates our central predictions.
%showing results that are inconsistent with a special mechanism theory of idioms, \fix{under which a high level of word co-occurrence is packaged together with non-canonical interpretation.}

We predict that the target phrases will score lower on conventionality than the matched phrases, since we expect these phrases to contain words with (often highly) unconventional meanings. We further predict that the target phrases will have higher contingency scores than the matched phrases, due to all of the target phrases being expressions that are frequently reused. Putting the two measures together, we expect idioms to fall at the intersection of low conventionality and high contingency, but not to show major discontinuities that qualitatively distinguish them from phrases that fall at other areas of intersection.

\subsection{Analysis 1: conventionality measure}\label{convanalysis}

We find that the target phrases have lower average conventionality scores than the matched phrases, with a difference of -1.654, with \textit{t}(145) = -5.829 and \textit{p} $<$ 0.001. This is consistent with idioms having unconventional word meanings.

\subsection{Analysis 2: contingency measure}\label{contingencymeasure}

We find that, averaged across contexts, the target phrases had higher contingency scores, with a difference in
value of 2.25 bits, with \textit{t}(159) = 8.807 and \textit{p} $<$ 0.001.

\begin{figure}[htb]
\begin{centering}
\includegraphics[scale=0.305]{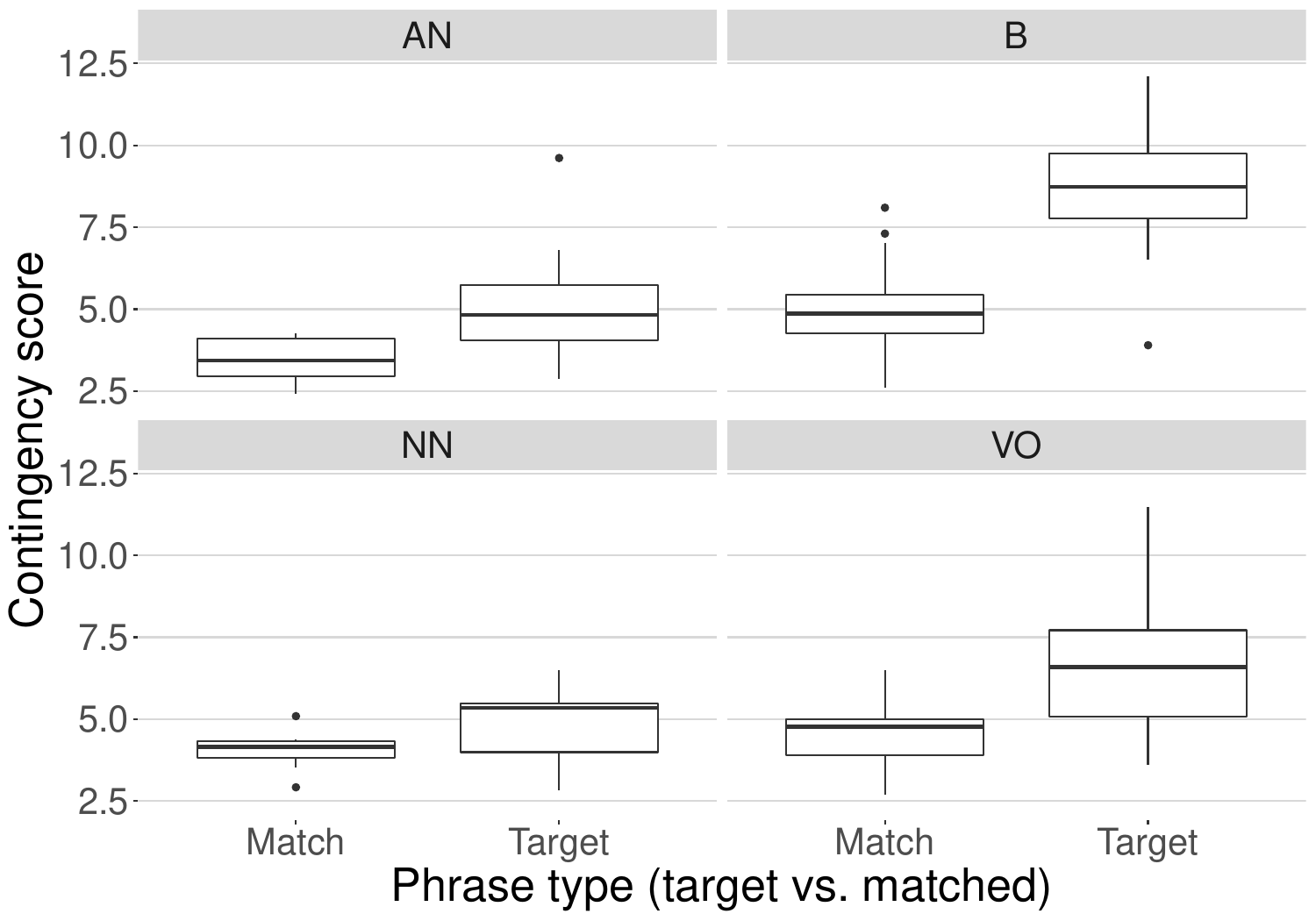}
	  \caption{\label{contingency30} Contingency of target and matched phrases, for phrases with at least 30 instances}
 \end{centering}
\end{figure}

Figure~\ref{contingency30} shows boxplots of the average contingency score for each phrase type. Since many of the target phrases only occurred in a handful of
sentences, we have excluded phrases for which the target or matched sets contain fewer
than 30 sentences.\footnote{This threshold was chosen to strike a balance between having enough instances contributing to the average score for each datapoint, and having a large enough sample of phrases. We considered thresholds at every multiple of 10 until we reached one that left at least 100 datapoints remaining.} For the most part, there
were fewer sentences containing the target phrase than there were
sentences containing only the head or only the non-head word in the relevant structural
position. This likely explains the greater variance among the target
phrases---the averages are based on fewer data points.

For all syntactic structures,
the median contingency score was higher for target phrases than matched phrases. The greatest differences were observed for verb object and
binomial phrases.

We fit another mixed effects model to test whether target idioms have higher contingency scores than matched phrases across syntactic classes (AN, B, NN, VO). The model predicts the contingencies for each instance of a phrase used in context, with the target-matched contrast and syntactic class as fixed effects, and random effects for the target-matched pairs.\footnote{\texttt{Cont$\sim$Target*Class$+$(1+Target|Idiom)}} We find that target phrases have significantly higher contingency scores than matched phrases (see Row 2 of Table~\ref{contingencymodelresults} of Appendix B).

%Given that many of the target phrases only occurred in a handful of sentences, Figure~\ref{contingency30} shows the same results, excluding phrases for which the target or matched sets contained less than 30 sentences. In this plot, all four structural phrase types have higher contingency for the target phrases than for the matched phrases. Among this subset, the target phrases had contingency values 2.57 higher than the matched phrases (\textit{t}(65) = -9.098, \textit{p} $<$ 0.001).

%\begin{figure}[htb]
%\begin{centering}
%\includegraphics[scale=0.3]{stickiness_30.pdf}
%\caption{\label{contingency30} Contingency of target and matched
%  phrases (for phrases with at least 30 instances)}
%	\end{centering}
%	\end{figure}

\subsection{Analysis 3: interaction and correlation of measures} \label{mainevent}

%\fix{Jackie's comment: "In terms of the content, the main thing that I'm thinking about right now is the space of phrases that we are considering in order to make the claim that conventionality and contingency are not correlated and phrases lie on a spectrum here. Theoretically speaking, are we interested in all phrases in a head-dependent relationship? If so, do we weight all such possible valid phrases equally, or by some notion of their frequency? I think this is important to think about, because it would help us evaluate the claims that we make about the lack of correlation + continuum arguments. What we've chosen to do is to select phrases that are matched with known idioms in a certain way. This works well for the analyses we make in Sections 5.1 and 5.2, but we might need to justify why this sampling choice is okay for the analyses in Section 5.3 wrt the underlying distribution of phrases we care about. In a way, it is actually surprising that we got the continuum effects, given that we had a particular way of sampling the matched phrases which is not just randomly sampling phrases in a corpus, right?"}

Here we show that idioms fall in the expected area of our two-dimensional space, with no evidence of correlation between the measures. Our results provide evidence against the notion of a special mechanism for idioms, whereby conventionality and contingency are expected to covary.

\begin{figure}[htb]
\begin{centering}
\includegraphics[scale=0.3]{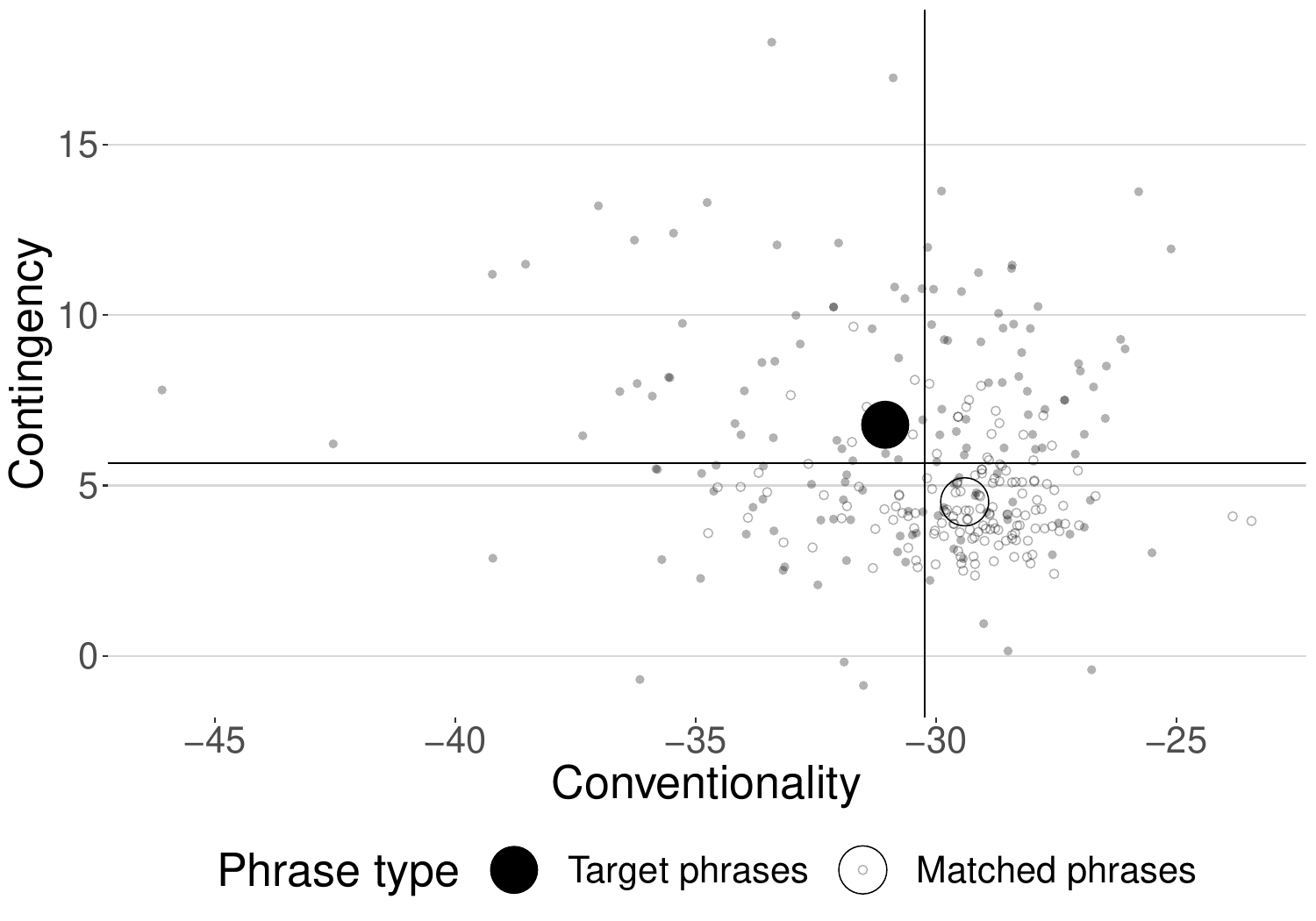}
	  \caption{\label{twobytwo_0} Contingency versus conventionality values of target and matched phrases. Large circles are average values of all target (black) and all matched (white) phrases.}
	\end{centering}
	\end{figure}

Recall the 2x2 matrix of contingency versus conventionality (Figure~\ref{table:2by2}), where idioms were expected to be in the top
left quadrant. Figure~\ref{twobytwo_0} shows our results. Since the conventionality scores
were for individual words, we averaged the scores of the head word and
the primary non-head word (i.e., the verb and the object for
verb object phrases, the adjective and the noun for adjective noun
phrases, the two nouns in noun noun phrases, and the two words of the
same category in binomial phrases). The plot shows the average values of the target and matched phrases. %only if the phrase and matched structure both occurred in at least 30 sentences in the BNC.

As discussed above, the target phrases came from lists of idioms in the literature, and thus include a mix of canonical idioms and (seemingly) compositional collocations. We predicted that the target phrases would be distributed between the top two quadrants, with obvious idioms on the
top left and collocations on the top right. As a sample, our results placed the following phrases in the top left quadrant: \textit{clear the air}, \textit{bread and butter}, \textit{nuts and bolts}, \textit{red tape}, and \textit{cut corners}. For each of these phrases, the idiomatic meaning cannot be derived by straightforwardly composing the meaning of the parts. In the top right quadrant (high conventionality, high contingency), we have \textit{more or less}, \textit{rise and fall}, \textit{back and forth}, and \textit{deliver the goods}. The bottom left quadrant was predicted to contain non-literal phrases whose words are not as strongly associated with one another as those in the most well-known idioms. The phrases in our dataset that fall into this quadrant include \textit{hard sell}, \textit{hit man}, and \textit{cold feet}. A list of which target phrases landed in each quadrant is given in Appendix D.

For the matched
phrases, we assumed that the majority were instances
of regular language use, so we predicted them to cluster in the bottom right quadrant. Our results are consistent with this prediction. The horizontal and vertical black lines on the plot were
placed at the mean values for each measure. Recall that our examples of ``regular language use’’ consist of head-dependent constructions that share one word with an existing idiom. Although obtaining the phrases in this way may have biased our sample of ``regular language use'' toward similarity with target phrases, the fact that we still see a clear difference between target and matched average values is all the more striking.

\begin{figure}[htb]
\begin{centering}
\includegraphics[scale=0.3]{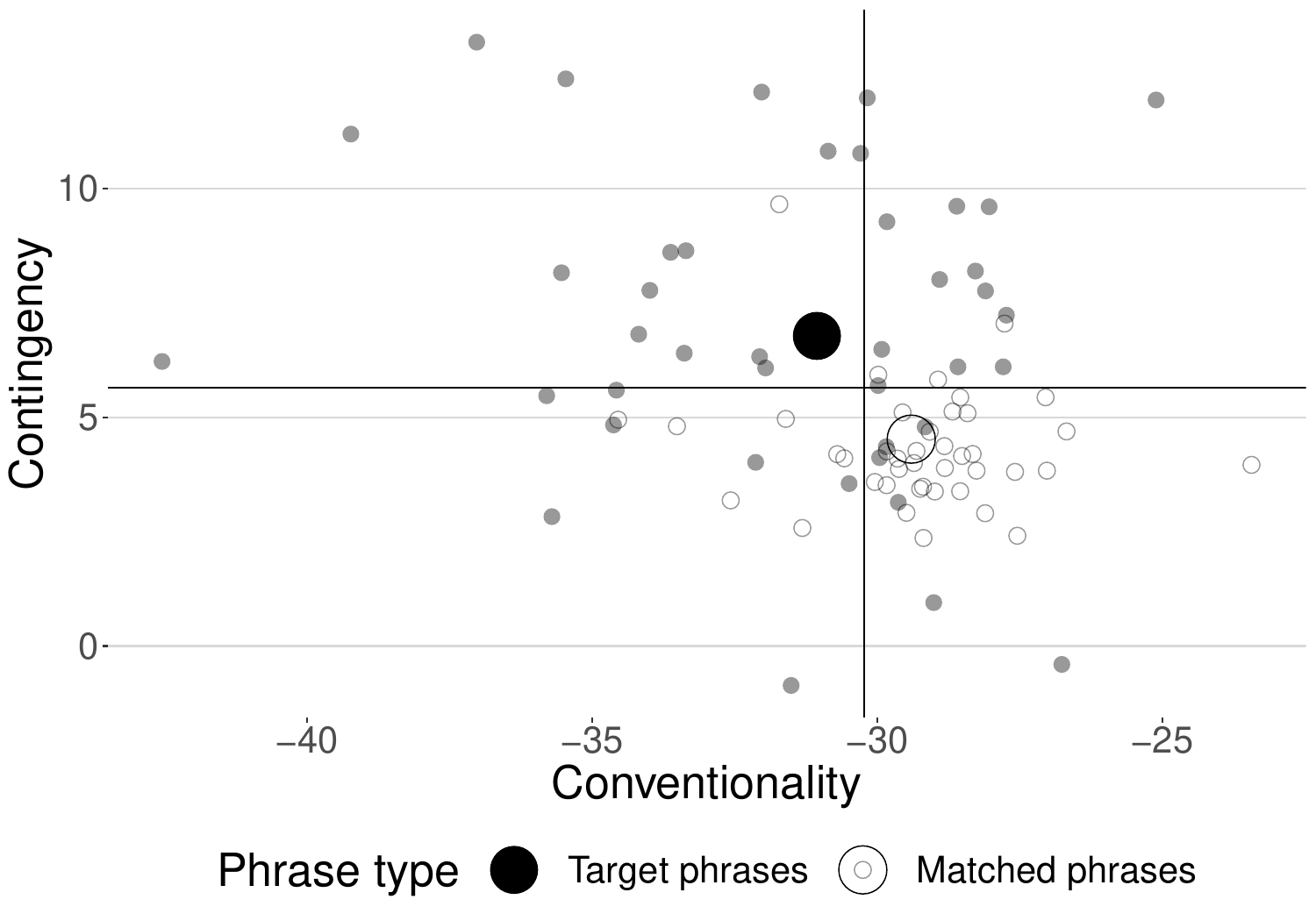}
	  \caption{\label{twobytwo_noncomp0} Contingency versus conventionality values of target and matched phrases (for target phrases rated as highly idiomatic). Large circles are average values of all target (black) and all matched (white) phrases.}
	\end{centering}
	\end{figure}

        Figure~\ref{twobytwo_noncomp0} shows only the target phrases
        that received a human annotation of 1 or 2 for head word
        literality---that is, the phrases judged to be most
        non-compositional. As expected, the average score for the
        target phrases moved more solidly into the idiom quadrant. %Note that, given the continuous nature of the measures, the division of language space into four quadrants is a simplification. There are, of course, intermediate areas that phrases can fall into.
        
We also found no evidence of correlation between
contingency and conventionality values among the entire set of phrases, target and matched (\textit{r}(312) =
-0.037, \textit{p} = 0.518), which is consistent with theories that treat the two properties as independent of each other.

\section{Asymmetries between heads and dependents}\label{asymmetriessec}

Our experiments revealed an unexpected but interesting
asymmetry between heads and their dependents. %We found that among the target phrases in particular, the head word behaved differently than the main non-head element (e.g., the verb versus the object in a verb-object idiom). This was especially evident in the conventionality scores, which we computed for each of the two content words individually.
Based on conventionality scores,
the head word of the target phrases was more conventional on average than
the primary non-head word. A two-sample t-test revealed that this
difference was significant ($t$ = 3.029, $df$ = 252.45, $p$ =
0.0027). The matched phrases did not show a significant difference between heads and non-heads ($t$ = 1.506, $df$ = 277.42, $p$ =
0.1332).

%To assess the relative contributions of the head and non-head words to a phrase's overall conventionality, a linear model was fit using the lmerTest package in R, with word conventionality as the dependent variable. The predictors were overall phrase compositionality and syntactic word (head versus non-head)

\begin{figure}[htb]
\begin{centering}
\includegraphics[scale=0.38]{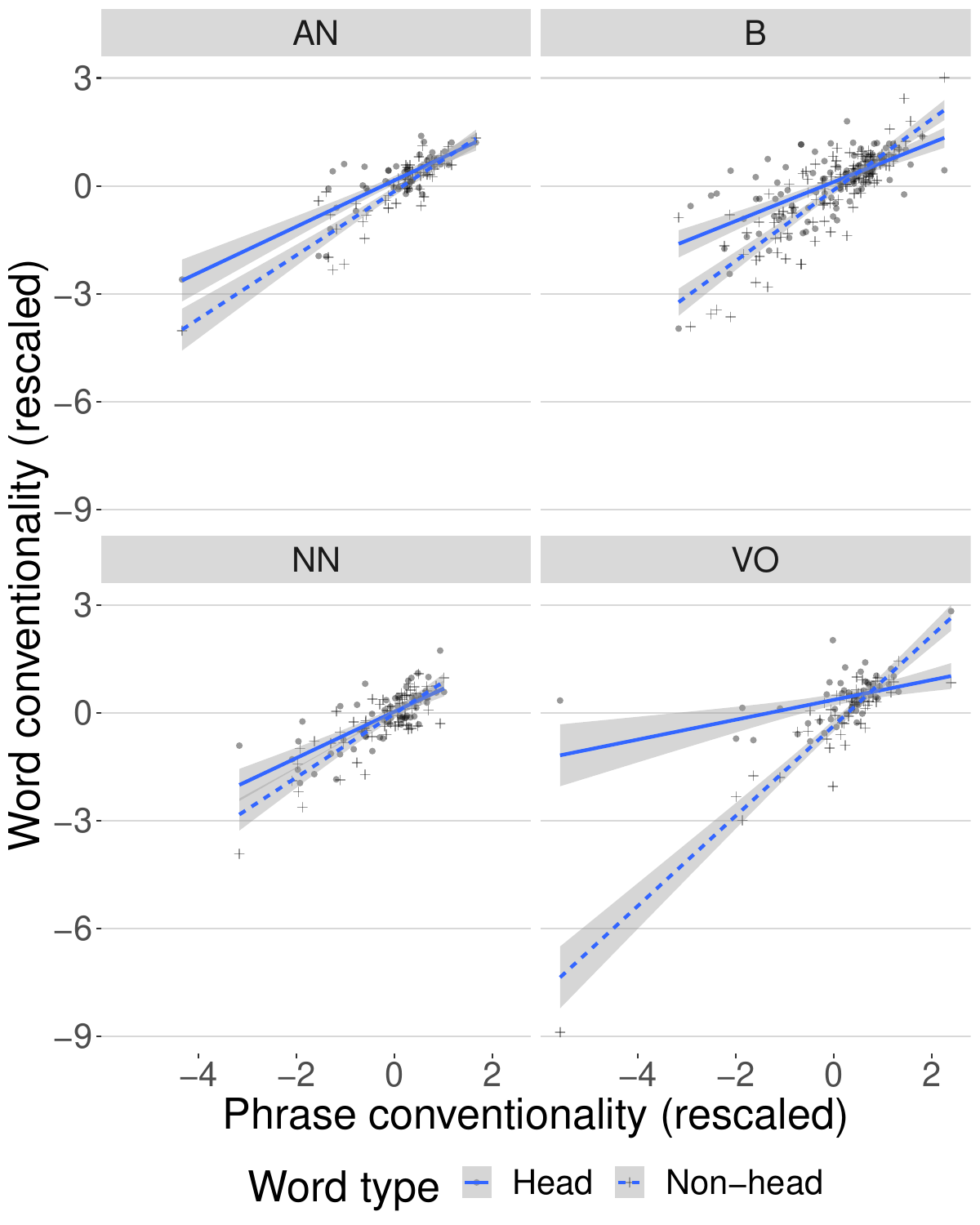}
	  \caption{\label{asymmetries} Change in head versus non-head conventionality scores as phrase conventionality increases, for all phrases (target and matched), separated by phrase type (adjective noun, binomial, noun noun, and verb object).}
	\end{centering}
	\end{figure}

Figure~\ref{asymmetries} presents the data in a different way, with target and matched phrases plotted together. The plots show that the variability in overall phrase conventionality, which helps to distinguish idioms and non-idioms, is largely driven by the dependent word (as indicated by the steeper slopes for the non-head effects). This interaction between phrase conventionality and head/non-head is significant (see Row 10 of Table~\ref{asymmetrymodelresults} of Appendix B).

In addition, Figure~\ref{asymmetries} illustrates that this discrepancy between heads and non-heads is largest for verb object phrases. We confirm this by fitting a linear model of word conventionality with predictors for phrase conventionality (average of the component words), head versus non-head word, and syntactic class, plus all interactions, using sum coding to compare factor levels of syntactic class.\footnote{\texttt{WordConv$\sim$PhraseConv*Class*Head}} The effect of headedness on conventionality scores is significantly greater for verb object phrases than the global effect of headedness (see Panel 4 of Figure~\ref{asymmetries}; Row 14 of Table~\ref{asymmetrymodelresults} of Appendix B). We raise the possibility that there is an additive effect of linear order, with conventionality decreasing from left to right through the phrase. For verb object phrases, the two effects go in the same direction, whereas for adjective noun and noun noun phrases, the linear order effect counteracts the headedness effect. We are not aware of any other theory positing the attribution of idiomatic meaning to incremental chunks in this way. Our results suggest that syntactic constituency alone is not enough to explain the observed patterns.

We note that there is disagreement in the literature about whether binomial phrases (which are coordinate structures) contain a head at all. Some proposals treat the first conjunct as the head
\citep[e.g.,][]{Rothstein:1991, Kayne:1994,gazdar.g:1981}, while others treat the conjunction as the head or claim that
there is no head \citep[e.g.,][]{Bloomfield:1933}. We find that
in the binomial case, the first conjunct patterns like the heads of
the other phrase types, though how much of this effect may be driven by linear order remains unclear. This may provide suggestive converging evidence for the first-conjunct-as-head theory, though further exploration of this idea is needed.

%We found a seemingly related effect when looking at the contingency scores. Recall that our contingency measure used XLNet to calculate joint probability estimates for particular phrases. We noted above that the conditional probabilities given by XLNet vary slightly depending on whether they are computed left-to-right or right-to-left. That is, the results shift slightly depending on which chain rule decomposition is used to compute the joint probability. Consider the example of \textit{spill the beans}. Computing left-to-right involves getting an XLNet value for the probability of \textit{spill}, given that one knows that \textit{beans} is in the object position. Computing right-to-left, on the other hand, involves getting the probability of \textit{beans} given \textit{spill}.

%For verb object and binomial target phrases, which are (arguably, in the case of binomials) head-initial, the contingency scores were higher when the left-to-right chain rule decomposition was used (by 0.76 for verb object phrases and 0.48 for binomials). For the head final target phrases (adjective noun and noun noun), the contingency scores were higher in the right-to-left direction (by 0.35 for both types of phrases). This indicates that the heads of these phrases are more predictable from the non-heads/dependents than the other way around.

%Together, these results suggest that the idiomaticity of the phrase does not come primarily from the head of the phrase, and that the non-head word is instead the stronger signal that one is dealing with an idiom.

\section{Related work}

%The individual measures described above have been instantiated in various ways in previous work on idiom detection.

Many idiom detection models build on insights about
unconventional meaning in metaphor. A number of approaches use
distributional models, such as
\citet{Kintsch:2000}, \citet{Utsumi:2011},
\citet{Sa-Pereira:2016}, and \citet{Shutovaetal:2012}, the latter of
which was one of the first to implement a fully unsupervised approach
for encoding relationships between words, their contexts, and
their dependencies. %\citet{green-etal-2011-multiword} demonstrate that idiom detection methods work better when they take into account syntactic structure rather than just surface statistics, primarily due to their ability to capture discontinuous expressions like phrasal verbs. In practice, this means relying on parsed corpora.
A related line of work aims to automatically
determine whether potentially idiomatic expressions are being used
idiomatically or literally, based on
contextual information
\citep{katz-giesbrecht-2006-automatic,Fazlyetal:2009,SporlederLi:2009,Pengetal:2014}. Our measure of conventionality is inspired by the insights of these models; as described in Section~\ref{convmeasure}, our measure uses differences in embeddings across contexts.

Meanwhile, approaches to collocation detection have taken a
probabilistic or information-theoretic approach that seeks to identify
collocations using word combination probabilities. PMI is a frequently-used
quantity for measuring co-occurrence probabilities
\citep[]{PMI,church-hanks-1990-word}. Other implementations
include selectional association \citep[][]{Resnik:1996}, symmetric
conditional probability \citep[][]{FerreiraPereira}, and
log-likelihood \citep[][]{dunning-1993-accurate,Daille:1996}. Like our study, most previous work on idiom and collocation detection focuses specifically on English.

While much of the literature in NLP recognizes that idioms share a cluster of properties, including semantic idiosyncrasy, syntactic inflexibility, and institutionalization
\citep[e.g.,][]{sag, FazlyStevenson:2006, Fazlyetal:2009}, our approach is
novel in attempting to characterize idioms along two orthogonal dimensions that correspond to specific proposals from the cognitive science literature. Our measures may offer a new avenue for tackling automatic idiom detection.

\section{Discussion \& Conclusion}

We investigated whether idioms could be characterized as occupying the intersection between contingency and conventionality, without needing to appeal to idiom-specific machinery that associates the storage of multi-word expressions with the property of unconventional meaning, as has been proposed in previous work.

%We validated that our measures corresponded to intuitions about language, and we showed that these measures concentrate idioms in the predicted region of the space. To compare the behavior of our conventionality measure with human behavior, we carried out an online rating experiment in which people rated the literalness of words and phrases in context. As predicted, higher literalness ratings correlated with higher conventionality scores given by our measure. As for the contingency measure, we separated the phrases in our corpus into idioms and collocations, on one hand, and non-idiom, non-collocations on the other. Again, as predicted, the idioms and collocations scored higher on our contingency measure.

When we plotted conventionality and contingency scores against each
other, we found that idioms fell, on average, in the area of low
conventionality and high contingency, as expected. Regular,
non-idiomatic phrases fell in the high conventionality, low
contingency area, also as expected. The lack of
correlation between the two measures provides support for theories
that divorce the notions of conventionality and
contingency. %Specifically, this lack of correlation suggests that the contingencyof the elements in a form is not dependent on the conventionality of those elements.

%We have thus provided evidence that idioms trend toward one end of the intersection between conventionality and contingency, but are not otherwise exceptional, contrary to theories that posit special machinery to handle idioms. %We have defined two measures, conventionality and contingency, which correspond to the notions of compositionality and storage, respectively, and shown that these measures are sufficient for characterizing idioms, but that they are not correlated and therefore provide no evidence of a dependence between the two mechanisms.
Our results suggest that idioms represent just one of the ways that
conventionality and contingency can interact, analogous to collocations or metaphor. We also presented the novel finding that the
locus of non-conventionality in idioms resides primarily in the dependent, rather than
the head, of the phrase, a result that merits further
study.

\section{Ethics statement}

This paper uses computational tools to argue for a theoretical position about idioms. Our idiom dataset was automatically generated from an existing corpus, and so did not involve data collection from human participants on our part. To validate our conventionality measure, we conducted an additional online experiment with crowdworkers on Amazon Mechanical Turk, for which we obtained REB approval. Details about the participants, recruitment, and consent process are given in Section~\ref{validation}.
We note that one limitation of this work is that it only investigates English idioms, potentially contributing to an over-focus on English in this domain.

\section*{Acknowledgments}

We thank Reuben Cohn-Gordon, Jacob Hoover, Alessandro Sordoni, and the Montreal Computational and Quantitative Linguistics Lab at McGill University for helpful feedback. We also gratefully acknowledge the support of the Natural Sciences and Engineering Research Council of Canada, the Fonds de Recherche du Qu\'{e}bec, and the Canada CIFAR AI Chairs Program.

\bibliography{Bibliography}
\bibliographystyle{acl_natbib}

%\end{document}

\newpage

\appendix

\section{}
\label{sec:appendix}
On the following page is a list of the target phrases in our corpus.

\begin{table}[htbp]\centering
\scalebox{0.85}{
\begin{tabular}{l c | l c}
\toprule
\textbf{Target phrase} & \textbf{Type} & \textbf{Target phrase} & \textbf{Type}\\ 
\midrule
deliver the goods & VO & swimming pool & NN \\
run the show &  VO & cash cow & NN  \\
rock the boat  & VO & foot soldier & NN   \\
call the shots & VO & attorney general & NN  \\
talk turkey  &  VO & hit list & NN  \\
cut corners & VO & soup kitchen & NN  \\
jump the gun  &  VO & bull market & NN  \\
have a ball & VO  & boot camp & NN \\
foot the bill  & VO & message board& NN   \\
break the mold & VO & gold mine & NN  \\
pull strings  &  VO & report card & NN  \\
mean business & VO & comfort food & NN  \\
raise hell  &  VO & pork barrel & NN \\
close ranks & VO & flower girl & NN  \\
strike a chord  & VO  & hit man & NN  \\
cry wolf & VO & blood money & NN  \\
lose ground  & VO  & cottage industry & NN  \\
make waves & VO & board game & NN  \\
clear the air  & VO & death wish & NN   \\
pay the piper & VO & word salad & NN  \\
spill the beans  & VO & altar boy & NN   \\
bite the dust & VO & bench warrant & NN  \\
saw logs & VO & time travel & NN  \\
lead the field & VO & love language & NN  \\
take the powder& VO & night owl & NN  \\
buy the farm &  VO & life blood& NN  \\
turn tail & VO & road rage & NN   \\
get the sack & VO & light house & NN  \\
hit the sack & VO & bid price & NN  \\
kick the bucket & VO & carrot cake & NN  \\
shoot the bull &  VO & command line & NN  \\
 &   & stag night & NN  \\
 &  & husband material & NN \\
&  \\
&  \\
&  \\
&  \\
&  \\
&  \\
 &   \\
&  \\
&  \\
&  \\
&  \\
 &   \\
&  \\
&  \\
&  \\
&  \\
&  \\
&  \\
&  \\

\bottomrule
\addlinespace[1ex]
\end{tabular}}
\end{table}

\begin{table}[htbp]\centering
\scalebox{0.85}{
\begin{tabular}{|l c | l c}
\toprule
\textbf{Target phrase} & \textbf{Type} & \textbf{Target phrase} & \textbf{Type}\\ 
\midrule
cold feet & AN & by and large & B \\
green light & AN & more or less & B  \\
red tape  & AN & bits and pieces & B   \\
black box & AN & up and down & B  \\
blue sky &  AN & rise and fall & B  \\
bright future & AN & sooner or later & B  \\
sour grape  & AN & rough and ready & B  \\
green room & AN  & far and wide & B \\
easy money  & AN & give and take & B  \\
last minute & AN & time and effort & B  \\
hard heart & AN & pro and con & B  \\
hot dog & AN & sick and tired & B   \\
raw talent  & AN & back and forth & B \\
hard labor & AN & day and night & B  \\
broken home  & AN  & wear and tear & B  \\
fat chance & AN & nut and bolt & B  \\
dirty joke  & AN  & tooth and nail & B  \\
happy hour & AN & on and off & B  \\
high time & AN & win or lose & B   \\
rich history & AN & food and shelter & B  \\
clean slate  & AN & odds and ends & B  \\
stiff competition & AN & in and out & B  \\
maiden voyage & AN & sticks and stones & B  \\
cold shoulder & AN & make or break & B  \\
clean energy & AN & part and parcel & B  \\
hard sell &  AN & loud and clear & B  \\
back pay & AN & cops and robbers & B   \\
deep pockets & AN & short and sweet & B  \\
broken promise & AN & safe and sound & B  \\
dead silence & AN & black and blue & B  \\
blind faith &  AN & toss and turn & B  \\
tight schedule & AN  & fair and square & B  \\
brutal honesty & AN & heads or tails & B \\
bright idea & AN & hearts and flowers & B\\
kind soul & AN & rest and relaxation & B\\
bruised ego & AN & flesh and bone & B\\
 &  & life and limb & B\\
 &  & checks and balances & B\\
 &  & fast and loose & B \\
 &  & high and dry & B\\
 &  & pots and pans & B\\
 &  & now or never & B\\
 &  & hugs and kisses & B\\
 &  & bread and butter & B\\
 &  & risk and reward & B\\
 &  & cloak and dagger & B\\
pins and needles & B & nickel and dime & B \\
sugar and spice & B & rhyme or reason & B \\
neat and tidy & B & leaps and bounds & B \\
step by step & B & live and learn & B \\
lost and found & B & peace and quiet & B \\
old and grey & B & song and dance & B \\

\bottomrule
\addlinespace[1ex]
\end{tabular}}
\end{table}

\newpage

\section{}
\label{sec:appendix2}

\begin{table}[htbp]\centering
\caption{Model results table with human literalness rating as the dependent variable, using \texttt{lmer}}
\tabcolsep=0.10cm
\begin{tabular}{l c c c l}
\toprule
\textbf{Coefficient} & \textbf{$\hat{\beta}$} & \textbf{$SE(\hat{\beta})$} & \textbf{$t$} & \textbf{$p$}\\ 
\midrule
\small Intercept    &    0.051   &   0.019   &  1.655  & 0.049 \\
\small Conv  &   0.185 &  0.050   &  3.725 &  $<$ 0.001 \\
\small Head(False)   &    0.015   & 0.014 & 1.050 & 0.147 \\
\small Conv:Head(False)   &    0.073   & 0.053 & 1.376 & 0.084 \\
\bottomrule
\addlinespace[1ex]
\multicolumn{3}{l}{$n$ = 4945}
\end{tabular}
\label{modelreport}
\end{table}

\begin{table}[htbp]\centering
\caption{Model results table for model described in Section~\ref{contingencymeasure}, with contingency score as the dependent variable, using \texttt{lmer}}
\tabcolsep=0.10cm
\begin{tabular}{l c c c l}
\toprule
\textbf{Coefficient} & \textbf{$\hat{\beta}$} & \textbf{$SE(\hat{\beta})$} & \textbf{$t$} & \textbf{$p$}\\ 
\midrule
\small Intercept    &    4.949   &   0.114   &  43.379  & $<$ 0.001 \\
\small Target(True)  &   1.253 &  0.165   &  7.587 &  $<$ 0.001 \\
\small Class(VO)   &    -0.195   & 0.200 & -0.975 & 0.165 \\
\small Class(AN)   &    -0.662   & 0.201 & -3.297 & $<$ 0.001 \\
\small Class(B)   &    1.796   & 0.179 & 10.045 & $<$ 0.001 \\
\small Target(True):   &    0.501   & 0.303 & 1.654 & 0.049 \\
Class(VO) & & & &
\\
\small Target(True):   &    -0.896   & 0.286 & -3.135 & $<$ 0.001 \\
Class(AN) & & & & 
\\
\small Target(True):   &    1.394   & 0.247 & 5.641 & $<$ 0.001
\\
Class(B) & & & & 
\\
\bottomrule
\addlinespace[1ex]
\multicolumn{3}{l}{$n$ = 99573}
\label{contingencymodelresults}
\end{tabular}
\end{table}
\newpage

\begin{table}[htbp]\centering
\caption{Model results table for model described in Section~\ref{asymmetriessec}, with conventionality score as the dependent variable}
\tabcolsep=0.10cm
\scalebox{0.9}{
\begin{tabular}{l c c c l}
\toprule
\textbf{Coefficient} & \textbf{$\hat{\beta}$} & \textbf{$SE(\hat{\beta})$} & \textbf{$t$} & \textbf{$p$}\\ 
\midrule
\small Intercept    &    0.163   &   0.035   &  4.614  & $<$ 0.001 \\
\small PhraseConv  &   0.526 &  0.036   &  14.453 &  $<$ 0.001 \\
\small Class(VO)   &    0.196   & 0.065 & 3.020 & 0.003 \\
\small Class(AN)   &    -0.135   & 0.063 & -2.153 & 0.032 \\
\small Class(B)   &    -0.010   & 0.064 & -0.150 & 0.881 \\
\small Head(False)  &   -0.326 &  0.050   &  -6.525 &  $<$ 0.001 \\
\small PhraseConv:Class(VO)   &    -0.250   & 0.062 & -4.043 & $<$ 0.001 \\
\small PhraseConv:Class(AN)   &    0.117   & 0.069 & 1.683 & 0.093 \\
\small PhraseConv:Class(B)   &    0.116   & 0.068 & 1.694 & 0.091 \\
\small PhraseConv:Head(False)   &    0.476   & 0.051 & 9.247 & $<$ 0.001 \\
\small Class(VO):Head(False)   &    -0.392   & 0.092 & -4.271 & $<$ 0.001 \\
\small Class(AN):Head(False)   &    0.271   & 0.089 & 3.044 & 0.002 \\
\small Class(B):Head(False)   &    0.019   & 0.091 & 0.212 & 0.832 \\
\small PhraseConv:Class(VO):   &    0.500   & 0.087 & 5.717 & $<$ 0.001 \\
Head(False) & & & &
\\
\small PhraseConv:Class(AN):  &    -0.233   & 0.098 & -2.380 & 0.018 \\
Head(False) & & & &
\\
\small PhraseConv:Class(B):   &    -0.232   & 0.097 & -2.396 & 0.017 \\
Head(False) & & & &
\\
\bottomrule
\addlinespace[1ex]
\multicolumn{3}{l}{$n$ = 584}
\label{asymmetrymodelresults}
\end{tabular}}
\end{table}

\newpage

\section{}
\label{sec:appendix3}

\begin{figure}[htb]
\begin{centering}
\includegraphics[scale=0.3]{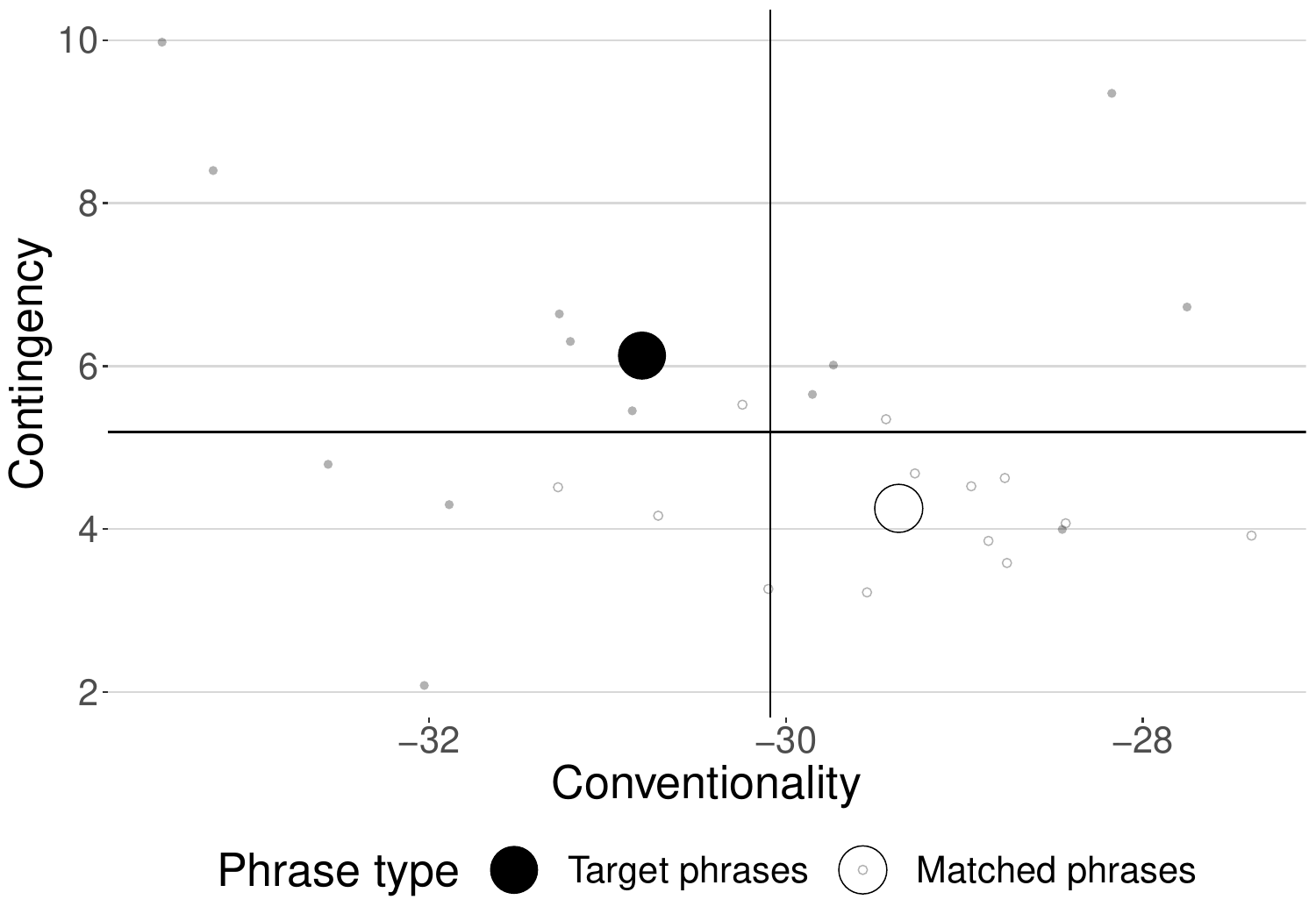}
	  \caption{\label{replication} Contingency and conventionality values of target and matched phrases. Large circles are average values of all target (black) and matched (white) phrases.}
 \end{centering}
\end{figure}

To confirm that our results are not simply an artifact of the dataset we used, we replicated the study on a second dataset, which is the set of phrases used in the idiom detection work of Fazly et al. (2009). We did not have any hand in choosing the phrases in this dataset, and it has very little overlap with our own. We once again fail to find evidence that the two dimensions of conventionality and contingency are correlated with one another in this set of phrases (\textit{r}(24) =
-0.276, \textit{p} = 0.172), and we see a similar spread of data across the four quadrants, shown in Figure~\ref{replication}.

\newpage

\section{}
\label{sec:appendix4}

Below is a list of the target phrases that landed in each of the quadrants in Figure~\ref{twobytwo_0}, for those phrases that occurred at least 30 times in the corpus.

\begin{table}[htbp]\centering
\scalebox{0.9}{
\begin{tabular}{c | c}
\toprule
\textbf{Top left} & \textbf{Top right}\\ 
\midrule
 black and blue & back and forth \\
 black box & bits and pieces \\
 bread and butter & boot camp \\
 by and large & bright future \\
 call the shots & deep pockets\\    
 checks and balances & deliver the goods \\
 clear the air & far and wide \\    
 cottage industry & food and shelter \\
 cut corners & heads or tails\\
 day and night & high and dry \\    
 foot soldier & more or less \\
 give and take & on and off \\    
 gold mine & part and parcel \\
 happy hour & pull strings \\
 have a ball & rise and fall \\
 high time & rock the boat \\
 in and out & run the show \\
 loud and clear & song and dance \\
 make or break & swimming pool \\
 nuts and bolts & up and down \\
 peace and quiet & \\
 red tape & \\
 safe and sound & \\
 sick and tired & \\
 soup kitchen & \\
 sour grapes & \\
 win or lose & \\

\bottomrule
\addlinespace[1ex]
\end{tabular}}
\end{table}

\begin{table}[htbp]\centering
\scalebox{0.9}{
\begin{tabular}{c | c}
\toprule
\textbf{Bottom left} & \textbf{Bottom right}\\ 
\midrule

 cold feet & blue sky \\
 green light & board game \\
 hard sell & bright idea \\
 hit man & get the sack \\
 hot dog & green room \\
 last minute & hit list \\
 lose ground & report card \\
 mean business & time and effort \\

\bottomrule
\end{tabular}}
\end{table}

\end{document}